\documentclass[11pt,a4paper]{article}
\usepackage[hyperref]{nlp4MusA}
\usepackage{times}
\usepackage{latexsym}
\usepackage{amsmath}
\usepackage{graphicx}
\usepackage{color}
\usepackage{url}
\usepackage{bm}

\aclfinalcopy % Uncomment this line for the final submission
\title{Generation of lyrics lines conditioned on music audio clips}

\author{Olga Vechtomova, Gaurav Sahu, Dhruv Kumar \\
  University of Waterloo \\
  \texttt{\{ovechtom, gsahu, d35kumar\}@uwaterloo.ca} \\
}

\date{}

\begin{document}
\maketitle
\begin{abstract}
We present a system for generating novel lyrics lines conditioned on  music audio. A bimodal neural network model learns to generate lines conditioned on any given short audio clip. The model consists of a spectrogram variational autoencoder (VAE) and a text VAE. Both automatic and human evaluations demonstrate effectiveness of our model in generating lines that have an emotional impact matching a given audio clip. The system is intended to serve as a creativity tool for songwriters.
\end{abstract}

\section{Introduction}
Creative text synthesized by neural text generative models can serve as inspiration for artists and songwriters when they work on song lyrics. Novel and unusual expressions and combinations of words in generated lines can spark an idea and inspire the songwriter to create original compositions. In contrast to systems that generate lyrics for an entire song, our system generates suggestions in the form of individual lyrics lines, and is intended to serve as a creativity tool for artists, rather than as a standalone songwriting AI system.

In a song, musical composition, instrumentation and lyrics act together to express the unique style of an artist, and create the intended emotional impact on the listener. Therefore, it is important that a lyrics generative model takes into account the music audio in addition to the textual content.

\begin{figure*}[!t]
	\centering \small
	\includegraphics[width=.85\linewidth]{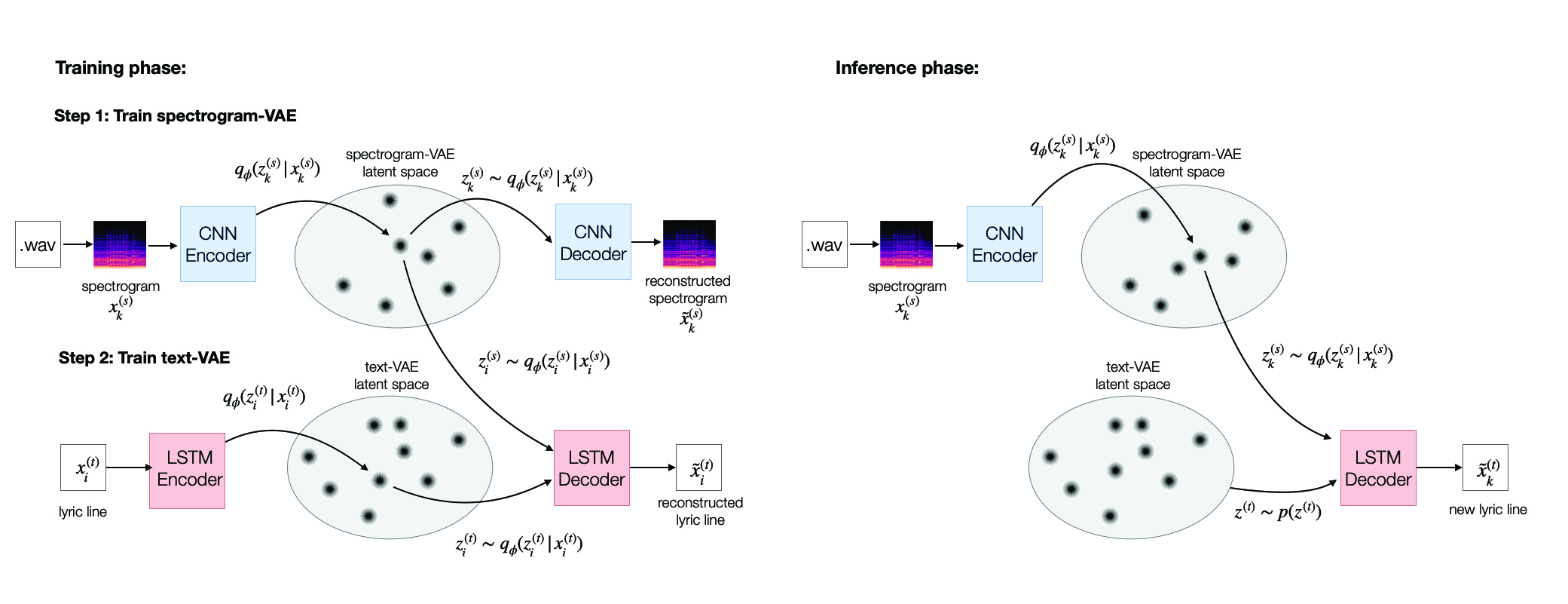}
	\caption{Music audio clip conditioned lyrics generation.}
	\vspace{-.5cm}
	\label{fig:audioClipVAE}
\end{figure*}

In this paper we describe a bimodal neural network model that uses music audio and text modalities for lyrics generation. The model (Figure~\ref{fig:audioClipVAE}) generates lines that are conditioned on a given music audio clip. The intended use is for an artist to play live or provide a pre-recorded audio clip to the system, which generates lines that match the musical style and have an emotional impact matching the given music piece.

The model uses the VAE architecture to learn latent representations of audio clips spectrograms. The learned latent representations from the spectrogram-VAE are then used to condition the decoder of the text-VAE that generates lyrics lines for a given music piece. Variational autoencoder lends itself very well for creative text generation applications, such as lyrics generation. It learns a latent variable model of the training dataset, and once trained any number of novel and original lines can be generated by sampling from the learned latent space.

Three main groups of approaches towards stylized text generation in natural language processing (NLP) include: (1) embedding-based techniques that capture the style information by real-valued vectors, and can be used to condition a language model~\cite{tikhonov2018guess} or concatenated with the input to a decoder ~\cite{persona}; (2) approaches that structure latent space to encode both style and content, and include Gaussian Mixture Model Variational Autoencoders (GMM-VAE)~\cite{shi2019fixing, wang2019topic}, Conditional Variational Autoencoders (CVAE)~\cite{yang2017generating}, and Adversarially Regularized Autoencoders (ARAE)~\cite{Li2020ComplementaryAC}; (3) approaches with multiple style-specific decoders~\cite{chen2019unsupervised}.

All of the above papers infer style from only one modality, text. Our work belongs to the first category of approaches: embedding based techniques, and is different from all of the above works in learning the style information from audio and text. Furthermore, previous embedding-based approaches use embeddings from a discrete vector space. This is the first work that uses a continuous latent variable learned by a spectrogram-VAE as a conditioning signal for the text-VAE.

A number of approaches have been proposed towards poetry generation, some focusing on rhyme and poetic meter \cite{zhang2014chinese}, while others on stylistic attributes \cite{tikhonov2018guess}. Cheng et al. \shortcite{cheng2018image} proposed image-inspired poetry generation. Yang et al. \shortcite{yang2017generating} generated poetry conditioned on specific keywords. Yu et al. \shortcite{yu2019deep} used audio data for lyrics retrieval. Watanabe et al. \shortcite{watanabe2018melody} developed a language model for lyrics generation using MIDI data. Vechtomova et al. \shortcite{vechtomova2018generating} generated author-stylized lyrics using audio-derived embeddings. To our knowledge this is the first work that uses audio-derived data to generate lyrics for a given music clip. 

The main contributions of this work are: (1) A probabilistic neural network model for generating lyrics lines matching the musical style of a given audio clip; (2) We demonstrate that continuous latent variables learned by a spectrogram-VAE can be effectively used to condition text generation; (3) Automatic and human evaluations show that the model can be effectively used to generate lyrics lines that are consistent with the emotional effect of a given music audio clip.

\section{Background: unconditioned text generation with VAE}
\label{sec:bg}
The variational autoencoder \cite{kingma2013auto} is a stochastic neural generative model that consists of an encoder-decoder architecture. The encoder transforms the input sequence of words $\bm x$ into the approximate posterior distribution $q_\phi(z|x)$ learned by optimizing parameters $\phi$ of the encoder. The decoder reconstructs $\bm x$ from the latent variable $\bm z$, sampled from $q_\phi(z|x)$. Both encoder and the decoder in our work are recurrent neural networks, specifically, Long Short Term Memory networks (LSTM). The reconstruction loss is the expected negative log-likelihood of data:

\vspace{-.5cm}
\begin{align}
J_\text{rec}(\phi, \theta, x)= -\sum_{t=1}^n \log p(x_t|\bm z, x_1\cdots x_{t-1})\label{eqn:rec_loss}
\end{align}
\vspace{-.5cm}

where $\phi$ and $\theta$ are parameters of the encoder and decoder, respectively. The overall VAE loss is

\vspace{-.5cm}
\begin{align}
J = J_\text{rec}(\phi, \theta, \bm x) + \text{KL}\!\left(\!q_\phi(\bm z|\bm x)\|p(\bm z)\!\right)\label{eqn:loss}
\end{align}
\vspace{-.5cm}

where the first term is the reconstruction loss and the second term is the KL-divergence between $\bm z$'s posterior and a prior distribution, which is typically set to standard normal $\mathcal{N}(\bm 0, \mathbf {I})$. 

\section{Approach}
\label{sec:audio_cg}

The audio clip conditioned generation model (Figure 2) consists of a spectrogram-VAE to learn a meaningful representation of the input spectrogram, and a text-VAE with an audio-conditioned decoder to generate lyrics.

In order to train the spectrogram-VAE, we first split the waveform audio of songs into small clips, and transform them into MEL spectrograms.
For this mode of generation, we generate data with two levels of audio-lyrics alignment: high-precision and low-precision, which are described in more detail in Section~\ref{sec:evaluation}.
The spectrogram-VAE follows a encoder-decoder architecture.
The encoder consists of four convolutional layers followed by a fully connected layer, and the decoder architechture is mirrored by using a fully-connected layer followed by four deconvolutional layers.  % any other details that can be mentioned here?
This model was trained for 100 epochs.  % confirm with Dhruv

We then feed all the spectrograms in the dataset to obtain their respective spectrogram embeddings. More precisely, we first obtain the $\boldsymbol{\mu}$ and $\boldsymbol{\sigma}$ for every data point, and then sample a latent vector from the learned posterior distribution using a random normal noise $\mathbf{\epsilon} \in \mathcal{N}(\bm 0, \mathbf{I})$.

The audio clip conditioned VAE, which has the same architecture as described in Section \ref{sec:bg}, is then trained to generate lyrics befitting the provided piece of music by concatenating the spectrogram embedding with the input to every step of the decoder.
The reconstruction loss is calculated as:

\vspace{-.8cm}
\begin{equation}
\begin{split}
    J_\text{rec}(\phi, \theta, {z_k}^{(s)}, x^{(t)})= -\sum_{i=1}^n \log p({x_i}^{(t)}|\bm {z_k}^{(t)},\\ 
    {z_k}^{(s)},
    {x_1}^{(t)}\cdots {x_{i-1}}^{(t)})
\end{split}
\label{eq:rec_loss_acg}
\end{equation}
\vspace{-.5cm}

where ${z_k}^{(s)}$, spectrogram embedding of the $k$-th data point.
At inference time, a latent vector ${z_k}^{(t)}$ sampled from the text-VAE's prior is concatenated with the corresponding spectrogram embedding ${z_k}^{(s)}$, and fed to the LSTM cell at every step of the text-VAE's decoder.

\section{Evaluation}
\label{sec:evaluation}

We collected a dataset of lyrics by seven Rock artists: David Bowie, Depeche Mode, Nine Inch Nails, Neil Young, Pearl Jam, Rush, and Doors. Each of them has a distinct musical and lyrical style, and a large catalogue of songs, spanning many years. We intentionally selected artists from the sub-genres of Rock as this models a real-world scenario, when a given songwriter might use the model to get influences from the genre congruent with their own work.

Since we do not have access to aligned data for these artists, we manually created a high-precision aligned dataset for two artists (Depeche Mode and Nine Inch Nails, 239 songs), and did an automatic coarse-grained alignment for the other five (518 songs). To create a manually aligned dataset, we annotated the original waveform files in Sonic Visualizer \cite{cannam2006sonic} by marking the time corresponding to the start of each lyrics line in the song.  The automatic alignment process consisted of splitting each song into 10-second segments and splitting lyrics lines into the same number of chunks, assigning each to the corresponding 10-second clip. In total, the training dataset consists of 18,210 lyrics lines and 14,670 spectrograms.

The goal of the developed model is to generate lyrics lines for an instrumental music piece. Our test set, therefore, only contains instrumental songs: 36 songs from an instrumental album "Ghosts I-IV" by Nine Inch Nails\footnote{Ghosts I-IV. Nine Inch Nails. Produced by: Atticus Ross, Alan Moulder, Trent Reznor. The Null Corporation. 2008. Released under Creative Commons (BY-NC-SA) license.} and eight instrumental songs from three other albums by two artists (Depeche Mode and Nine Inch Nails). Each song was split into 10-second clips, which were then converted into spectrograms (807 in total).

First we evaluate the quality of the latent space learned by the spectrogram-VAE. For every spectrogram in the test set, we computed pairwise cosine similarity between its embedding ${z}^{(s)}$ and the embedding of every spectrogram in the training and test set. We then calculated the proportion of clips in the top 50 and 100 that (a) are part of the same song, (b) are part of the same album, and (c) belong to the same artist. The results (Table~\ref{tab:spec-VAE}) indicate that large proportions of clips most similar to a given clip are by the same artist and from the same album. This demonstrates that spectrogram-VAE learns representations of an artist's unique musical style.
\vspace{-.2cm}
\begin{table}[!htpb]
\small
	\centering
%	\resizebox{\linewidth}{!}{
		\begin{tabular}{ c | c | c | c}
			\hline
			\textbf{Top-n} & \textbf{same song} & \textbf{same album} & \textbf{same artist}\\
			\hline
			n=50 & 0.1707 & 0.4998 & 0.7293 \\
			\hline
			n=100 & 0.0988 & 0.4462 & 0.7067 \\
			\hline
		\end{tabular}
	\caption{Clustering effect in the spectrogram-VAE latent space. }\vspace{-.2cm}
	\label{tab:spec-VAE}
\end{table}

 We divided songs in the test set into two categories: ``intense'' and ``calm''. The peak dB differences between tracks in these two categories are statistically significant (t-test, p\textless0.05). A spectrogram for each 10-second clip was used to generate 100 lines according to the method described in Section \ref{sec:audio_cg}. The songs in these two categories evoke different emotions and we expect that the lexicon in these two categories of generated lines will be different, but more similar among songs within the same category. 

Automatic evaluation of generated lines conditioned on an instrumental audio clip is difficult, since there is no reference ground-truth line that we can compare the generated line to. For this reason, we cannot use n-gram overlap based measures, such as BLEU. Secondly, style-adherence metrics, such as classification accuracy w.r.t. a certain class, e.g. style, in the dataset are inapplicable, since there is no categorical class variable here.  

We calculated KL divergence values for words in each song. KL divergence measures the relative entropy between two probability distributions. It was defined in information theory \cite{losee1990science} and was formulated as a word ranking measure in \cite{carpineto2001information}. 
Given word $w$ in the generated corpus $G_k$ conditioned on clip $k$ and generated corpus $N$ conditioned on all other clips in the test set, KL divergence is calculated as
$\text{word-KL}(w) = p_{Gk}(w) \cdot
\text{log}(p_{Gk}(w) / p_N(w))$.

We then calculated rank-biased overlap scores (RBO) to measure pairwise similarity of word-KL ranked lists corresponding to each pair of songs. RBO \cite{webber2010similarity} is a metric developed in Information Retrieval for evaluating ranked search results overlap, and handles non-conjoint ranked lists. The RBO score falls in the range [0,1] where 0 indicates a disjoint list and 1 - identical. 

\begin{figure}[!h]
	\centering \small
	\includegraphics[width=1\linewidth]{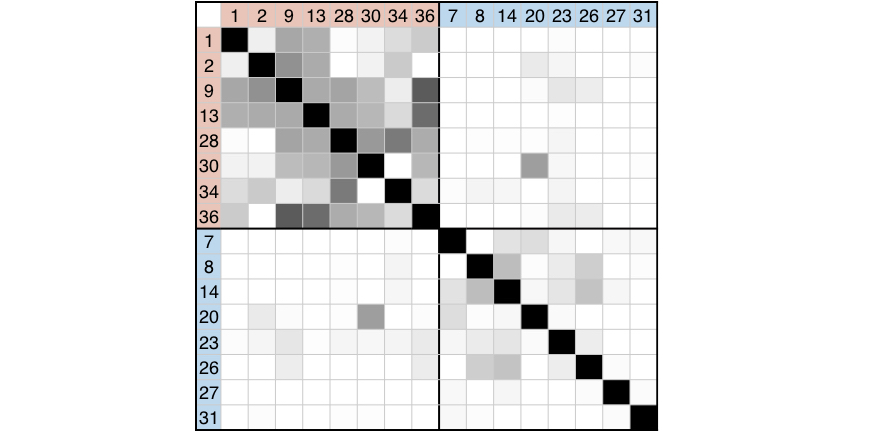}
	\caption{Rank-biased overlap (RBO) between the KL-divergence ranked lists of generated words for songs in ``Ghosts I-IV'' album. Pink-highlighted songs are calm, and blue - intense. The darker grey cells indicate higher overlap. The row/column headings correspond to the numbers in the song titles (e.g. 2 is for ``2 Ghosts I'')
	.}
	\vspace{-.2cm}
	\label{fig:rbo-loud-quiet}
\end{figure}

Figure \ref{fig:rbo-loud-quiet} shows that ``calm'' songs have higher RBO values with other songs in the same category, indicating similar generated word distributions, and low RBO values w.r.t. ``intense'' songs. RBO values for lines generated conditioned on ``intense'' songs are not as high, suggesting that they have less word overlap. This is likely because there are more songs in this category in the training set with widely different lyrics, therefore the model may be picking up more subtle musical differences, which make it correspondingly generate lyrics that have lyrical influences from different songs.
	\vspace{-.2cm}

\begin{figure}[h!]
	\centering \small
	\includegraphics[width=1\linewidth]{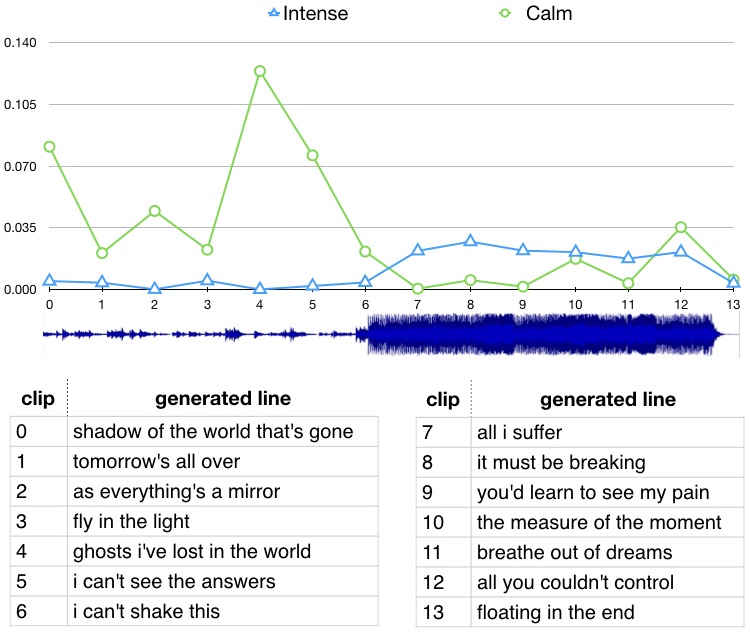}
	\caption{Rank-biased overlap (RBO) values for 10-second clips of ``12 Ghosts II'' song and examples of lines generated for each clip.}
	\vspace{-.5cm}
	\label{fig:waveform-rbo}
\end{figure}

The difference between word distributions is also evident at the audio clip level. Figure~\ref{fig:waveform-rbo} shows RBO values for each 10-second clip in the song ``12 Ghosts II''\footnote{Demos of generated lines are available at:\\
\url{https://sites.google.com/view/nlp4musa-submission/home}}. The x-axis is the timeline. We first calculated word-KL for every clip w.r.t. all other clips in the test set. Then pairwise RBO was computed between the given clip's ranked word list and the ranked word lists for ``intense'', and ``calm'' generated corpora, respectively. The original song's waveform is given for reference, showing correlation with the change in the lexicon being generated for calm and intense sections of the track.
	\vspace{-.2cm}

We have also conducted a human evaluation. Six participants (3 female and 3 male), none of whom are members of the research team, were asked to listen to ten instrumental music clips from the test set. For each clip they were given two lists of 100 generated lines. One list was generated conditioned on the given clip in either ``calm'' or ``intense'' category, the other list was generated based on a clip from the opposite category. The participants were asked to select the list that they thought was generated based on the given clip. The average accuracy was 78.3\% (sd=9.8), which shows that participants were able to detect emotional and semantic congruence between lines and a piece of instrumental music.

\section{Conclusions}
We developed a bimodal neural network model, which generates lyrics lines conditioned on an instrumental audio clip. The evaluation shows that the model generates different lines for audio clips from ``calm'' songs compared to ``intense'' songs. Also, songs in the ``calm'' category are lexically more similar to each other than to the songs in the ``intense'' category. A human evaluation shows that the model learned meaningful associations between the semantics of lyrics and the musical characteristics of audio clips captured in spectrograms.

\bibliography{nlp4MusA}
\bibliographystyle{nlp4MusA_natbib}

\end{document}